\documentclass[runningheads]{llncs}
\usepackage{booktabs}
\usepackage{multirow}
\usepackage{graphicx}
\usepackage{amsmath}
\usepackage{amssymb}
\begin{document}
\title{AutoAM: An End-To-End Neural Model for Automatic and Universal Argument Mining}
% \title{Contribution Title\thanks{Supported by organization x.}}
%
\titlerunning{AutoAM}

% \author{Anonymous Author(s)\inst{1}}
\author{Lang Cao\inst{1}}
\authorrunning{L. Cao}
\institute{University of Illinois Urbana Champaign, United States \\
\email{langcao2@illinois.edu}}
\maketitle              % typeset the header of the contribution
\begin{abstract}
Argument mining is to analyze argument structure and extract important argument information from unstructured text. An argument mining system can help people automatically gain causal and logical information behind the text. As argumentative corpus gradually increases, like more people begin to argue and debate on social media, argument mining from them is becoming increasingly critical. However, argument mining is still a big challenge in natural language tasks due to its difficulty, and relative techniques are not mature. For example, research on non-tree argument mining needs to be done more. Most works just focus on extracting tree structure argument information. Moreover, current methods cannot accurately describe and capture argument relations and do not predict their types. In this paper, we propose a novel neural model called AutoAM to solve these problems. We first introduce the argument component attention mechanism in our model. It can capture the relevant information between argument components, so our model can better perform argument mining. Our model is a universal end-to-end framework, which can analyze argument structure without constraints like tree structure and complete three subtasks of argument mining in one model. The experiment results show that our model outperforms the existing works on several metrics in two public datasets.

\keywords{Argument Mining \and Information Extraction \and Natural Language Processing}
\end{abstract}
\section{Introduction}
Argument mining (AM) is a technique for analyzing argument structure and extracting important argument information from unstructured text, which has gained popularity in recent years \cite{lawrence_argument_2020}. An argument mining system can help people automatically gain causal and logical information behind the text. The argument mining techniques benefit plenty of many fields, like legal \cite{walker-etal-2018-evidence}, public opinions \cite{PARK18.679}, finance, etc. Argument mining is beneficial to human society, but there is still much room for development. Argument mining consists of several tasks and has a variety of different paradigms \cite {lawrence_argument_2020}. In this paper, we focus on the most common argument structure of monologue. It is an argumentative text from one side, not an argument from two sides. The microscopic structure of argumentation is the primary emphasis of the monologue argument structure, which primarily draws out the internal relations of reasoning.

\begin{figure}[htbp]
\centerline{\resizebox{6cm}{!}{\includegraphics{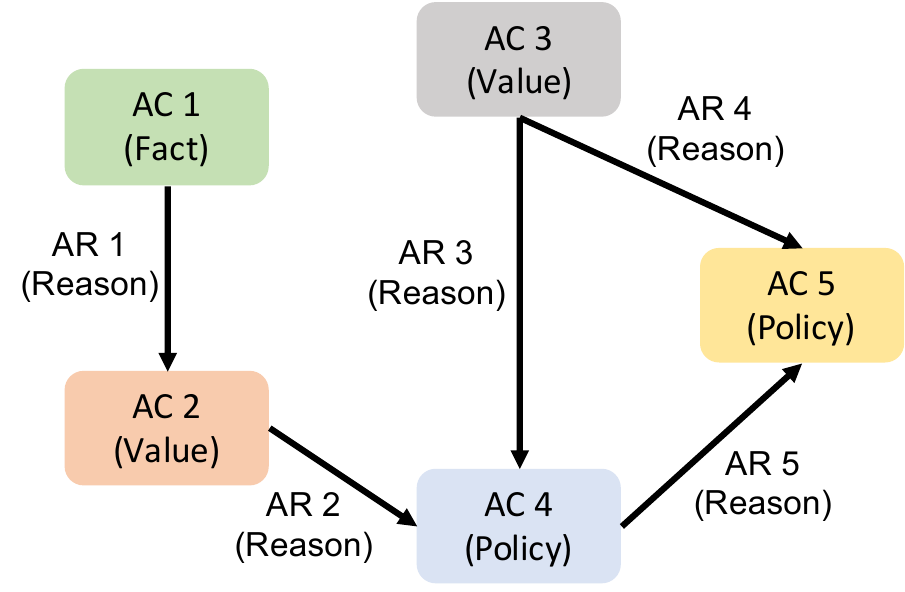}}}
\caption{An example of argument mining result after extraction in the CDCP dataset \cite{PARK18.679}. It forms an argument graph. In this graph, every node AC represents an argument component. Fact, Value, and Policy are three types of ACs. Every edge AR denotes argument relation, and Reason is one type of AR.}
\label{fig:example}
\end{figure}

In this setting, an argumentative paragraph can be viewed as an argument graph. An argument graph can efficiently describe and reflect logical information and reasoning paths behind the text. An example of AM result after extraction is shown in Figure \ref{fig:example}. The two important elements in an argument graph are the argument component (AC) and the argument relation (AR). ACs are nodes in this graph, and ARs are edges. The goal of an AM system is to construct this argument graph from unstructured text automatically. The process of the AM system definition we use is as following steps:

\begin{enumerate}
    \item Argument Component Identification (ACI): Given an argumentative paragraph, AM systems will detect ACs from it and separate this text.
    \item Argument Component Type Classification (ACTC): AM systems will determine the types of these ACs.
    \item Argument Relation Identification (ARI): AM systems will identify the existence of a relationship between any ACs.
    \item Argument Relation Type Classification (ARTC): AM systems will determine the type of ARs, which are the existing relations between ACs.
\end{enumerate}

Subtask 1) is a token classification task, which is also a named entity recognition task. This task has a large amount of research work on it. Most of the previous argument mining works \cite{potash-etal-2017-heres} \cite{kuribayashi-etal-2019-empirical} \cite{chakrabarty-etal-2019-ampersand} assume that the subtask 1) argument component identification has been completed, which is the argument component has been identified and can be obtained from the argumentative text. Therefore, the emphasis of argument mining research is placed on other subtasks. Following previous works, we also make such assumptions in this paper. On this basis, we design an end-to-end model to complete ACTC, ARI, and ARTC subtasks simultaneously.

ARI and ARTC are the hardest parts of the whole argument mining process. An AR is represented by two ACs. It is difficult to represent AR precisely and capture this relation. Most ACs pairs do not have a relationship at all, which leads to a serious sample imbalance problem. Among the whole process, ARI and ARTC are parts of ACI and ACTC, so the performance of these tasks will be influenced. Due to these reasons, many previous works give up and ignore the classification of ARs. Besides, much research imposes some argument constraints to do argument mining. In most normal cases, they assume argument information is a tree structure, and they can use the characteristic of the tree to extract information. Tree structure argument information is common in an argumentative essay. However, argument information with no constraints is more normal in the real world, like a huge amount of corpus on social media. This information is just like the general argument graphs mentioned before and needs to be extracted in good quality.

In this paper, we solve the above problems with a novel model called \textbf{AutoAM} (the abbreviation of \textbf{Auto}mactic and Universal \textbf{A}rgument \textbf{M}ining Model). This is an efficient and
accurate argument mining model to complete the entire argument mining process. This model does not rely on domain-specific corpus and does not need to formulate special syntactic constraints, etc., to construct argument graphs from argumentative text. To improve the performance of non-tree structured argument mining, we first introduce the argument component attention mechanism (\textbf{ArguAtten}) in this model, which can better capture the relevant information of argument components in an argumentative paragraph. It benefits the overall performance of argument mining. We use a distance matrix to add the key distance feature to represent ARs. A stratified learning rate is also a critical strategy in the model to balance multi-task learning. To the best of our knowledge, we are the first to propose an end-to-end universal AM model without structure constraints to complete argument mining. Meanwhile, we combine our novelty and some successful experience to achieve the state of the art in two public datasets.

In summary, our contributions are as follows:
\begin{itemize}
    \item We propose a novel model \textbf{AutoAM} for argument mining which can efficiently solve argument mining in all kinds of the argumentative corpus.
    \item We introduce \textbf{ArguAtten} (argument component attention mechanism) to better capture the relation between argument components and improve overall argument mining performance.
    \item We conduct extensive experiments on two public datasets and demonstrate that our method substantially outperforms the existing works. The experiment results show that the model proposed in this paper achieves the best results to date in several metrics. Especially, there is a great improvement over the previous studies in the tasks of ARI (argument relation identification) and ARTC (argument relation type classification).
\end{itemize}

\section{Related Work}
Since argument mining was first proposed \cite{inproceedings}, much research has been conducted on it. At first, people used rule-based or some traditional machine learning methods. With the help of deep learning, people begin to get good performance on several tasks and start to focus on non-tree structured argument mining. We discuss related work following the development of AM.

\subsection{Early Argument Mining}
The assumption that the argument structure could be seen as a tree or forest structure was made in the majority of earlier work, which made it simpler to tackle the problem because various tree-based methods with structural restrictions could be used. In the early stage of the development of argument mining, people usually use rule-based structural constraints and traditional machine learning methods to conduct argumentative mining. In 2007, Moens et al. \cite{inproceedings} conducted the first argument mining research on legal texts in the legal field, while Kwon et al. \cite{10.5555/1248460.1248473} also conducted relevant research on commentary texts in another field. However, the former only identified the content of the argument and did not classify the argument components. Although the latter one further completed the classification of argument components, it still did not extract the relationship between argument components, and could not explore the argument structure in the text. It only completed part of the process of argument mining.

\subsection{Tree Structured Argument Mining with Machine Learning}
According to the argumentation paradigm theory of Van Eemeren et al. \cite{eemeren_grootendorst_2003}, Palau and Moens \cite{inproceedings_mr} modeled the argument information in legal texts as a tree structure and used the hand-made Context-Free Grammar (CFG) to parse and identify the argument structure of the tree structure. This method is less general and requires different context-free grammars to be formulated for different structural constraints of argument. By the Stab and Gurevych \cite{stab-gurevych-2014-annotating} \cite{stab-gurevych-2017-parsing} tree structure of persuasive Essay (Persuasive Essay, PE) dataset has been in argument mining has been applied in many studies and practices. In this dataset, Persing and Ng \cite{persing-ng-2016-end} and Stab and Gurevych \cite{stab-gurevych-2017-parsing} used the Integer Linear Programming (ILP) framework to jointly predict the types of argument components and argument relations. Several structural constraints are defined to ensure a tree structure. The arg-micro text (MT) dataset created by Peldszus \cite{peldszus-2014-towards} is another tree-structured dataset. In studies using this dataset, decoding techniques based on tree structure are frequently used, such as Minimum Spanning tree (MST) \cite{peldszus-stede-2015-joint}, and ILP \cite{article_as}.

\subsection{Neural Network Model in Argument Mining}
With the popularity of deep learning, neural network models have been applied to various natural language processing tasks. For deep learning methods based on neural networks, Eger et al. \cite{eger-etal-2017-neural} studied argument mining as a sequence labeling problem that relies on parsing multiple neural networks. Potash et al. \cite{potash-etal-2017-heres} used sequence-to-sequence pointer network \cite{NIPS2015_29921001} in the field of argument mining and identified the different types of argument components and the presence of argument relations using the output of the encoder and decoder, respectively. Kuribayashi et al. \cite{kuribayashi-etal-2019-empirical} developed a span representation-based argumentation structure parsing model that employed ELMo \cite{peters-etal-2018-deep} to derive representations for ACs.

\subsection{Recent Non-Tree Structured Argument Mining}
Recently, more works have focused on the argument mining of non-tree structures. The US Consumer Debt Collection Practices (CDCP) dataset \cite{niculae-etal-2017-argument} \cite{PARK18.679} greatly promotes the development of non-tree structured argument mining. The argument structures contained in this dataset are non-tree structures. On this dataset, Niculae et al. \cite{niculae-etal-2017-argument} carry out a structured learning method based on a factor graph. This method can also handle the tree structure of datasets. It can also be used in the PE dataset, but the factor diagram needs a specific design according to the different types of the argument structure. Galassi et al. \cite{galassi-etal-2018-argumentative} used the residual network on the CDCP dataset. Mor IO et al. \cite{morio-etal-2020-towards} developed an argument mining model, which uses a task-specific parameterized module to encode argument. In this model, there is also a bi-affine attention module \cite{dozat-manning-2018-simpler} to capture the argument. Recently, Jianzhu Bao et al. \cite{bao-etal-2021-neural}tried to solve both tree structure argument and non-tree structure argument by introducing the transformation-based dependency analysis method \cite{chen-manning-2014-fast} \cite{gomez-rodriguez-etal-2018-global}. This work gained relatively good performance on the CDCP dataset but did not complete the ARTC task in one model and did not show the experiment results of ARTC.

However, these methods either do not cover the argument mining process with a good performance or impose a variety of argument constraints. There is no end-to-end model for automatic and universal argument mining before. Thus, we solve all the problems above in this paper.

\section{Methodology}
\begin{figure}[htbp]
\centerline{\resizebox{6cm}{!}{\includegraphics{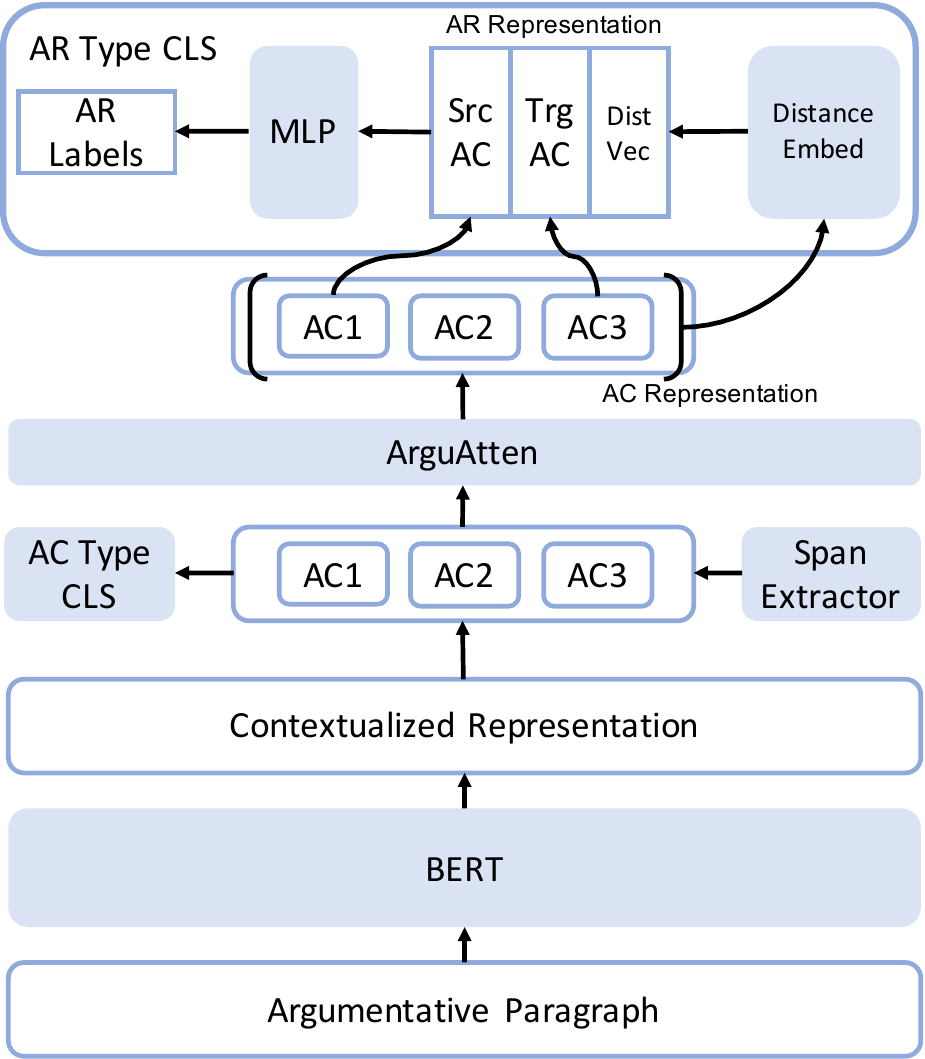}}}
\caption{The framework of our proposed model called AutoAM.}
\label{fig:AutoAM}
\end{figure}

As shown in Figure \ref{fig:AutoAM}, we propose a new model called AutoAM. This model adopts the joint learning approach. It uses one model to simultaneously learn the ACTC, ARI, and ARTC three subtasks in argument mining. For the argument component extraction, the main task is to classify the argument component type, and the argument component identification task has been completed by default on both the PE and the CDCP datasets. For argument relation extraction, the model regards ARI and ARTC as one task. The model classifies the relationship between the argument components by a classifier and then gives different prediction results for two tasks by post-processing prediction labels.

\subsection{Task Formulation}
The input data contains two parts: a) A set of $n$ argumentative text $T = \{T_1, T_2, ..., T_n\}$, b) for the $i$th argumentative text, there are $m$ argument component spans  $S = \{S_1, S_2, ..., S_m\}$, where every span marks the start and end scope of each AC $S_i = (start_i, end_i)$. Our aim is to train an argument mining model and use it to get output data: a) types of $m$ ACs provided in the input data $ACs = \{AC_1, AC_2, ..., AC_m\} $, b) $k$ existing ARs $ARs = \{AR_1, AR_2, ..., AR_k\}$ and their types, where $AR_i = (AC_a \rightarrow AC_b)$.

\subsection{Argument Component Extraction}
By default, the argument component identification task has been completed. The input of the whole model is an argumentative text and a list of positional spans corresponding to each argument component $S_i = (start_i, end_i)$.

We input argumentative text $T$ into pre-trained language models (PLMs) to get contextualized representations $H \in \mathbb{R}^{m \times d_b}$, where $d_b$ is the dimension of the last hidden state from PLMs. Therefore, we represent argumentative text as $H = (h_1, h_2, ..., h_m)$, where $h_i$ denotes the $i$th token contextualized representation.

We separate argument components from the paragraph using argument component spans $S$. In the PE dataset, the argument components do not appear continuously. We use mean pooling to get the representation of each argument component. Specifically, the $i$ argument component can be represented as:
\begin{equation}
AC_i=\frac{1}{end_i-start_i+1}\sum^{end_i}_{j=start_i} h_i,
\end{equation}
where $AC_i\in\mathbb{R}^{d_b}$. Therefore, all argument components in the argumentative text can be represented as $ACs = (AC_1, AC_2, ..., AC_n)$. For each argument component, we input it into AC Type Classifier $MLP_a$ in order. This classifier contains a multi-layer perceptron. A Softmax layer is after it. The probability of every type of argument component can be get by:
\begin{equation}
p(y_i|AC_i)=Softmax(MLP_a(AC_i)),
\end{equation}
where $y_i$ represent the predicted labels of the $i$th argument component. We get the final predicted label of its argument component as:
\begin{equation}
\hat{y}_i=Argmax(p(y_i|AC_i)).
\end{equation}

\subsection{Argument Relation Extraction}
This model views ARI and ARTC as having the same task and distinguish them by post-processing predictions. We classify every argument component pair $(AC_i \rightarrow AC_j)$. Argument component pairs are different of $(AC_i \rightarrow AC_j)$ and $(AC_j \rightarrow AC_i)$. We add a label, `none' here. `none' represents that there is no relation of $AC_i\rightarrow AC_j$.

In the argument relation extraction part, we use the enumeration method. We utilize output results from the ACTC step. We combine two argument components and input them into AR Type Classifier to get the predicted output.

First, the model uses ArguAtten (Argument Component Attention mechanism) to enhance the semantic representation of argument components. The self-attention mechanism is first proposed in the Transformer \cite{NIPS2017_3f5ee243}. The core of this mechanism is the ability to capture how each element in a sequence relates to the other elements, i.e., how much attention each of the other elements pays to that element. When the self-attention mechanism is applied to natural language processing tasks, it can often capture the interrelationship of all lexical elements in a sentence and better strengthen the contextual semantic representation. In the task of argument mining, all argument components in an argumentative text also meet this characteristic. The basic task of argument mining is to construct an argument graph containing nodes and edges, where nodes are argument components and edges are argument relations. Before the argument relation extraction task, the self-attention mechanism of argument components can be used to capture the mutual attention of argument components. It means that it can better consider and capture the argument information of the full text. This mechanism is conducive to argument relations extraction and the construction of an argument graph. We define ArguAtten as:
\begin{equation}
ArguAtten(Q,K,V)=Softmax(\frac{QK^T}{\sqrt{d_k}})\times V,
\end{equation}
where $Q$, $K$, $V$ are got by multiplying ACs with $W_Q$, $W_K$, $W_V$. They are three parameter matrices $W_Q, W_K, W_V \in \mathbb{R}^{d_b \times d_k}$, and $d_k$ is the dimension of attention layer. Besides, we also use ResNet and layer normalization (LN) after the attention layer to avoid gradient explosion:
\begin{equation}
ResNetOut=LN(ACs + ArgutAtten(ACs)).
\end{equation}
Through the self-attention of argument components, we obtain a better contextualized representation of argument components and then begin to construct argument pairs to perform argument relation extraction.

We consider that the relative distance between two argument components has a decisive influence on the type of argument relations between them. By observing the dataset, we can find that there is usually no argument relation between the two argument components, which are relatively far apart. It can significantly help the model to classify the argument relation types. Therefore, we incorporate this feature into the representation of argument relations. At first, the distance vector is introduced, and the specific definition is shown as:
\begin{equation}
V_{dist}=(i-j)\times W_{dist},
\end{equation}
where $(i-j)$ represents a relative distance, it can be positive or negative. $W_{dist}\in\mathbb{R}^{1\times d_{dist}}$ is a distance transformation matrix, and it can transform a distance scalar to a distance vector. $d_{dist}$ is the length of the distance vector.

For each argument relation, it comes from the source argument component (Src AC), the target argument component (Trg AC), and the distance vector (Dist Vec). We concatenate them to get the representation of an argument relation as:
\begin{equation}
AR_{i,j}=[AC_i, AC_j, V_{dist}],
\end{equation}
where $AR_{i,j}\in\mathbb{R}^{d_b\times2+d_{dist}}$, $d_{dist}$ is the length of distance vector.

Therefore, argument relations in an argumentative text can be represented as $ARs=(AC_{1,2}, AC_{1,3}, ..., AC_{n,n-1},)$, contains $n\times(n-1)$ potential argument relations in total. We do not consider self-relation like $AR=(AC_i \rightarrow AC_i)$.

For each potential argument relation, we separately and sequentially input them into the AR Type Classifier $MLP_b$. The classifier uses a Multi-Layer Perceptron (MLP) containing a hidden layer of 512 dimensions. The output of the last layer of the Multi-layer Perceptron is followed by a Softmax layer to obtain the probability of an argument relationship in each possible type label, as shown in:
\begin{equation}
p(y_{i,j}|AR_{i,j})=Softmax(MLP_b(AR_{i,j})),
\end{equation}
where $y_{i,j}$ denotes the predicted label of the argument relation from the $i$th argument component to the $j$th argument component. The final predicted labels are:
\begin{equation}
\hat{y}_{i,j}=Argmax(p(y_{i,j}|AR_{i,j})).
\end{equation}

To get the predicted labels of ARI and ARTC, we post-processed the prediction of the model. The existence of an argument relation in the ARI task is defined as:
% \begin{equation}
% \hat{y}_{ARI}=\left\{
% \begin{cases}
% 0,\quad if\quad \hat{y}_{AR} = 0 \\z
% 1,\quad if\quad \hat{y}_{AR} \neq 0 \\
% \end{cases}
% \right. ,
% \end{equation}

\begin{equation}
\hat{y}_{ARI} =
\begin{cases}
0 & \text{if } \hat{y}_{AR} = 0 \\
1 & \text{if } \hat{y}_{AR} \neq 0
\end{cases}
\end{equation}
where $\hat{y}_{AR}$ is the predicted label from the model output.

When we gain the type of an existing argument relation in the ARTC task, we assign the probability of `none' to zero and select the other label with the higher probability. They are represented as:
\begin{equation}
\hat{y}_{ARTC}=Argmax(p(y_{AR}|AR_{i,j})), \quad y^{none}=0,
\end{equation}
where $y^{none}$ is the model output of the label `none'.

\subsection{Loss Function Design}
This model jointly learns the argument component extraction and the argument relation extraction. By combining these two tasks, the training objective and loss function of the final model is obtained as:
\begin{equation}
L(\theta)=\sum_i log(p(y_i|AC_i))+\sum_{i,j}p(y_{i,j}|AR_{i,j})+\frac{\lambda}{2}||\theta||^2,
\end{equation}
where $\theta$ represents all the parameters in the model, and $\lambda$ represents the coefficient of L2 regularization. According to the loss function, the parameters in the model are updated repeatedly until the model achieves better performance results to complete the model training.

\section{Experiments}
\subsection{Datasets}
We evaluate our proposed model on two public datasets: Persuasive Essays (PE) \cite{stab-gurevych-2017-parsing} and Consumer Debt Collection Practices (CDCP) \cite{niculae-etal-2017-argument}.

The PE dataset only has tree structure argument information. It has three types of ACs: \textit{Major-Claim}, \textit{Claim}, and \textit{Premise}, and two types of AR: \textit{support} and \textit{attack}.

The CDCP dataset has general structure argument information, not limited to a tree structure. It is different from the PE dataset and is more difficult. The argument information in this dataset is more similar to the real world. There are five types of ACs (propositions): \textit{Reference}, \textit{Fact}, \textit{Testimony}, \textit{Value}, and \textit{Policy}. Between these ACs, there are two types of ARs: \textit{reason} and \textit{evidence}.

We both use the original train-test split of two datasets to conduct experiments.

\subsection{Setups}
In the model training, roberta-base \cite{DBLP:journals/corr/abs-1907-11692} was used to fine-tune, and AdamW optimizer \cite{DBLP:journals/corr/abs-1711-05101} was used to optimize the parameters of the model during the training. We apply a stratified learning rate to obtain a better semantic representation of BERT context and downstream task effect. The stratified learning rate is important in this task because this multi-task learning is complex and have three subtasks. The ARI and ARTC need a relatively bigger learning rate to learn the data better. The initial learning rate of the BERT layer is set as 2e-5. The learning rate of the AC extraction module and the AR extraction module is set as 2e-4 and 2e-3, respectively. After BERT output, the Dropout Rate \cite{JMLR:v15:srivastava14a} is set to 0.2. The maximum sequence length of a single piece of data is 512. We cut off ACs and ARs in the over-length text. The batch size in each training step is set to 16 in the CDCP dataset and 2 in the PE dataset. The reason is that there are more ACs in one argumentative text from the PE dataset than in the CDCP dataset.

In training, we set an early stop strategy with 5 epochs. We set the minimum training epochs as 15 to wait for the model to become stable. We use $Macro F1_{ARI}$ as monitoring indicators in our early stop strategy. That is because AR extraction is our main improvement direction. Furthermore, the ARI is between the ACTC and the ARTC, so we can better balance the three tasks' performance in the multi-task learning scenario.

The code implementation of our model is mainly written using PyTorch \cite{https://doi.org/10.48550/arxiv.1912.01703} library, and the pre-trained model is loaded using Transformers \cite{https://doi.org/10.48550/arxiv.1910.03771} library. In addition, model training and testing were conducted on one NVIDIA GeForce RTX 3090.

\subsection{Compared Methods}
We compare our model with several baselines to evaluate the performance:
\begin{itemize}
    \item \textbf{Joint-ILP} \cite{stab-gurevych-2017-parsing} uses Integer Linear Programming (ILP) to extract ACs and ARs. We compare our model with it in the PE dataset.
    \item \textbf{St-SVM-full} \cite{niculae-etal-2017-argument} uses full factor graph and structured SVM to do argument mining. We compare our model with it in both the PE and the CDCP datasets.
    \item \textbf{Joint-PN} \cite{potash-etal-2017-heres} employs a Pointer Network with an attention mechanism to extract argument information. We compare our model with it in the PE dataset.
    \item \textbf{Span-LSTM} \cite{kuribayashi-etal-2019-empirical} use LSTM-based span representation with ELMo to perform argument mining. We compare our model with it in the PE dataset.
    \item \textbf{Deep-Res-LG} \cite{galassi-etal-2018-argumentative} uses Residual Neural Network on AM tasks. We compare our model with it in the CDCP dataset.
    \item \textbf{TSP-PLBA} \cite{morio-etal-2020-towards} introduces task-specific parameterization and bi-affine attention to AM tasks. We compare our model with it in the CDCP dataset.
    \item \textbf{BERT-Trans} \cite{bao-etal-2021-neural} use transformation-based dependency analysis method to solve AM problems. We compare our model with it in both the PE and the CDCP datasets. It is also the state of the art on two datasets.
\end{itemize}

\subsection{Performance Comparison}

\begin{table*}[]
\centering
\resizebox{12cm}{!}{
\begin{tabular}{@{}l|cccccc|ccc|ccc|c@{}}
\toprule
\multirow{2}{*}{Methods} & \multicolumn{6}{c|}{ACTC} & \multicolumn{3}{c|}{ARI} & \multicolumn{3}{c|}{ARTC} & \multirow{2}{*}{AVG} \\
 & Macro & Value & Policy & Testi. & Fact & Refer. & Macro & Rel. & Non-rel. & Macro & Reason & Evidence &  \\ \midrule
St-SVM-strict & 73.2 & 76.4 & 76.8 & 71.5 & 41.3 & 100.0 & - & 26.7 & - & - & - & - & - \\
Deep-Res-LG & 65.3 & 72.2 & 74.4 & 72.9 & 40.3 & 66.7 & - & 29.3 & - & 15.1 & 30.2 & 0.0 & - \\
TSP-PLBA & 78.9 & - & - & - & - & - & - & 34.0 & - & - & - & 18.7 & - \\
BERT-Trans & 82.5 & 83.2 & 86.3 & 84.9 & 58.3 & 100.0 & 67.8 & 37.3 & 98.3 & - & - & - & - \\ \midrule
\textbf{AutoAM (Ours)} & \textbf{84.6} & \textbf{85.0} & \textbf{86.8} & \textbf{86.1} & \textbf{65.9} & \textbf{100.0} & \textbf{68.4} & \textbf{38.5} & \textbf{98.4} & \textbf{71.3} & \textbf{98.1} & \textbf{44.4} & \textbf{74.8} \\ \bottomrule
\end{tabular}
}
\caption{The results of comparison experiments on the CDCP dataset. All numbers in the table are f1 scores (\%). The best scores are in bold. `-' represents that the original paper does not report.}
\label{tab:cdcp_results}
\end{table*}

\begin{table}[]
\centering
\resizebox{12cm}{!}{
\begin{tabular}{@{}l|cccc|ccc|ccc|c@{}}
\toprule
\multirow{2}{*}{Methods} & \multicolumn{4}{c|}{ACTC} & \multicolumn{3}{c|}{ARI} & \multicolumn{3}{c|}{ARTC} & \multirow{2}{*}{AVG} \\
 & Macro & MC & Claim & Premise & Macro & Rel. & Non-rel. & Macro & Support & Attack &  \\ \midrule
Joint-ILP & 82.6 & 89.1 & 68.2 & 90.3 & 75.1 & 58.5 & 91.8 & 68.0 & 94.7 & 41.3 & 75.2 \\
St-SVM-strict & 77.6 & 78.2 & 64.5 & 90.2 & - & 60.1 & - & - & - & - & - \\
Joint-PN & 84.9 & 89.4 & 73.2 & 92.1 & 76.7 & 60.8 & 92.5 & - & - & - & - \\
Span-LSTM & 85.7 & 91.6 & 73.3 & 92.1 & 80.7 & 68.8 & 93.7 & \textit{79.0} & \textit{96.8} & \textbf{61.1} & 81.8 \\
BERT-Trans & \textit{88.4} & \textbf{93.2} & \textit{78.8} & \textit{93.1} & \textbf{82.5} & \textbf{70.6} & \textit{94.3} & \textbf{81.0} & - & - & \textbf{83.4} \\ \midrule
\textbf{AutoAM (Ours)} & \textbf{88.7} & \textit{91.9} & \textbf{80.3} & \textbf{93.9} & \textit{81.6} & \textit{65.8} & \textbf{98.5} & 75.4 & \textbf{97.6} & \textit{53.2} & \textit{81.9} \\ \bottomrule
\end{tabular}
}
\caption{The results of comparison experiments on the PE dataset. All numbers in the table are f1 scores (\%). The best results are in bold. The second best results are in italics. `-' represents that the original paper does not report.}
\label{tab:pe_results}
\end{table}

The evaluation results are summarized in Table \ref{tab:cdcp_results} and Table \ref{tab:pe_results}. In both tables, `-' indicates that the original paper does not measure the performance of this metric for its model. The best results are in bold, and the second-best results are in italics.

On the CDCP dataset, we can see our model achieves the best performance on all metrics in ACTC, ARI, and ARTC tasks. We are the first to complete all the tasks and get ideal results on the CDCP dataset. Our model outperforms the state of the art with an improvement of 2.1 in ACTC and 0.6 in ARI. The method BERT-Trans does not perform ARTC with other tasks at the same time, and it does not report results of ARTC, maybe due to unsatisfactory performance. In particular, compared with the previous work, we have greatly improved the task performance of ARTC and achieved ideal results. % 372%

On the PE dataset, our model also gets ideal performance. However, we get the second-best scores in several metrics. The first reason is that the PE dataset is tree-structured, so many previous work impose some structure constraints. Their models incorporate more information, and our model assumes they are general argument graphs in contrast. Another reason is that the models BERT-Trans, Span-LSTM, and Joint-PN combine extra features to represent ACs, like paragraph types, BoW, position embedding, etc. This information will change in the different corpus, and we want to build an end-to-end universal model. For example, there is no paragraph type information in the CDCP dataset. Therefore, we do not use them in our model. Even if our model does not take these factors into account, we achieve similar results to the state of the art.

\subsection{Ablation Study}

\begin{table}[]
\centering
\begin{tabular}{@{}l|cc|cc|cc|cc@{}}
\toprule
\multirow{2}{*}{Methods} & \multicolumn{2}{c|}{ACTC} & \multicolumn{2}{c|}{ARI} & \multicolumn{2}{c|}{ARTC} & \multicolumn{2}{c}{AVG} \\
 & Macro & $\bigtriangledown$ & Macro & $\bigtriangledown$ & Macro & $\bigtriangledown$ & Macro & $\bigtriangledown$ \\ \midrule
\textbf{AutoAM (Ours)} & \textbf{84.6} & - & 68.4 & - & \textbf{71.3} & - & \textbf{74.8} & - \\ \midrule
w/o stratified learning rate & 76.9 & -7.7 & 66.2 & -2.2 & 49.3 & -22.0 & 64.2 & -10.6 \\
w/o ArguAtten & 82.9 & -1.7 & \textbf{69.8} & \textbf{+1.4} & 57.7 & -13.6 & 70.1 & -4.7 \\
w/o Distance Matrix & 82.9 & -1.7 & 62.9 & -5.5 & 59.3 & -12.0 & 68.3 & -6.5 \\ \bottomrule
\end{tabular}
\caption{The results of ablation experiments on the CDCP dataset. All numbers in the table are f1 scores (\%). The best results are in bold.}
\label{tab:ablation_results}
\end{table}

The ablation study results are summarized in Table \ref{tab:ablation_results}. We conduct an ablation study on the CDCP dataset to see the impact of key modules in our model. It can be observed that the stratified learning rate is the most critical in this model. It verifies the viewpoint that multi-task learning is complex in this model and ARs extraction module needs a bigger learning rate to perform well. We can see ArguAtten improve the ACTC and ARTC performance by 1.7 and 13.6. However, the ARI matric decreases a little bit. Even though the numbers are small, we think that the reason is the interrelationship between ACs has little impact on the prediction of ARs' existence. ArguAtten mainly plays an effect in predicting the type of ARs. From this table, we can also find that the distance matrix brings the important distance feature to AR representation with an overall improvement of 6.5.

\section{Conclusion and Future Work}
In this paper, we propose a novel method for argument mining and first introduce the argument component attention mechanism. This is the first end-to-end argument mining model that can extract argument information without any structured constraints and get argument relations of good quality. In the model, ArguAtten can better capture the correlation information of argument components in an argumentative paragraph so as to better explore the argumentative relationship. Our experiment results show that our method achieves the state of the art.
In the future, we will continue to explore designing a better model to describe and capture elements and relationships in argument graphs.

% \subsubsection*{Acknowledgements.}

\bibliographystyle{splncs04}
\bibliography{references}




% LNCS Homepage, \url{http://www.springer.com/lncs}. Last accessed 4
% \end{thebibliography}
\end{document}